\pdfoutput=1
\documentclass[11pt]{article}

\usepackage[preprint]{acl}

\usepackage{times}
\usepackage{latexsym}

\usepackage[utf8]{inputenc}

\usepackage{microtype}

\usepackage{inconsolata}

\usepackage{subcaption}
\usepackage{colortbl}
\usepackage{multirow}
\usepackage{graphicx}
\usepackage{amssymb}
\usepackage{booktabs}
\usepackage{hyperref}

\definecolor{codegreen}{rgb}{0,0.6,0}
\definecolor{codegray}{rgb}{0.5,0.5,0.5}
\definecolor{codepurple}{rgb}{0.58,0,0.82}
\definecolor{backcolour}{rgb}{0.95,0.95,0.92}
\newcommand{\highlight}[1]{\color{blue}{#1}}
\newcommand{\onedot}{.}

\usepackage{listings} % For lstlisting
\usepackage{color} % For defining colors
\usepackage{xspace} % For defining colors
\usepackage{fancyvrb}
\usepackage{fvextra} % Extended features for fancyvrb
\usepackage[most]{tcolorbox}

\lstdefinestyle{mystyle}{
    columns=fullflexible,
    backgroundcolor=\color{backcolour},   
    commentstyle=\color{codegreen},
    numberstyle=\tiny\color{codegray},
    stringstyle=\color{codepurple},
    basicstyle=\footnotesize,
    breakatwhitespace=false,         
    breaklines=true,  
    breakindent=0pt,
    captionpos=b,                    
    keepspaces=true,  
    numbersep=0pt,                  
    showspaces=false,                
    showstringspaces=false,
    showtabs=true,                  
    tabsize=10,
    escapeinside={@}{@}, % Define escape sequence for LaTeX inside listings
    moredelim=**[is][\highlight]{`}{`}, % Highlight words with `
}
\usepackage{natbib}
\usepackage[T1]{fontenc}
\usepackage[most]{tcolorbox}
\usepackage{lmodern} 
\usepackage{lipsum}
\definecolor{plotly_green}{HTML}{60CCA2}
\definecolor{plotly_red}{HTML}{DD6D56}
\definecolor{plotly_blue}{HTML}{656EF1}

\title{Reasoning Paths with Reference Objects Elicit Quantitative Spatial Reasoning in Large Vision-Language Models}
\author{
    \textbf{Yuan-Hong Liao\textsuperscript{1}},
    \textbf{Rafid Mahmood\textsuperscript{2,3}},
    \textbf{Sanja Fidler\textsuperscript{1,2}},
    \textbf{David Acuna\textsuperscript{2}}
    \\
    \\
    \textsuperscript{1} University of Toronto, Vector Institute
    \textsuperscript{2} NVIDIA
    \textsuperscript{3} University of Ottawa
    \\
    \small{
    \textbf{Correspondence:} \href{mailto:andrew@cs.toronto.edu}{andrew@cs.toronto.edu}
  } 
}

\newcommand{\AL}[1]{{\color{magenta}{[Andrew: #1]}}}

\newcommand{\DA}[1]{{\color{red}{[David: #1]}}}
\newcommand{\RRM}[1]{{\color{green}{[Rafid: #1]}}}
\newcommand{\arxiv}[1]{{\color{blue}{#1}}}
\def\gemini{Gemini 1.5 Pro\xspace}
\def\geminiflash{Gemini 1.5 Flash\xspace}
\def\gpto{GPT-4o\xspace}
\def\gptv{GPT-4V\xspace}

\def\scannetqas{Q-Spatial-ScanNet\xspace}
\def\inhouseqas{Q-Spatial++\xspace}
\def\benchmark{Q-Spatial Bench\xspace}
\def\OurMethod{\texttt{SpatialPrompt}\xspace}
\def\OurMethodSimple{\texttt{SpatialPrompt}-\texttt{Single}\xspace}
\def\OurMethodComplex{\texttt{SpatialPrompt}-\texttt{Steps}\xspace}

\def\eg{\emph{e.g}\onedot} 

\def\ie{\emph{i.e}\onedot}

\begin{document}
\maketitle
\begin{abstract}

Despite recent advances demonstrating vision-language models' (VLMs) abilities to describe complex relationships in images using natural language, their capability to quantitatively reason about object sizes and distances remains underexplored. 
In this work, we introduce a manually annotated benchmark, \benchmark, with 271 questions across five categories designed for quantitative spatial reasoning and systematically investigate the performance of state-of-the-art VLMs on this task.
% 
% In this work, we systematically investigate current SoTA VLMs using a manually annotated benchmark of 241 questions across five categories specifically designed for quantitative spatial reasoning.
% 
Our analysis reveals that reasoning about distances between objects is particularly challenging for SoTA VLMs; however, some VLMs significantly outperform others, with an over 40-point gap between the two best performing models. 
We also make the surprising observation that the success rate of the top-performing VLM increases by 19 points when a reasoning path using a reference object emerges naturally in the response. 
Inspired by this observation, we develop a zero-shot prompting technique, \OurMethod, that encourages VLMs to answer quantitative spatial questions using reference objects as visual cues. 
By instructing VLMs to use reference objects in their reasoning paths via \OurMethod, \gemini, \geminiflash, and \gptv improve their success rates by over 40, 20, and 30 points, respectively. 
We emphasize that these significant improvements are obtained without needing more data, model architectural modifications, or fine-tuning.\footnote{Project website: \url{https://andrewliao11.github.io/spatial_prompt}}
\end{abstract}

\section{Introduction}

% \AL{We need to mention somewhere that we are doing QAs, which might be more relevant to NLP guys.} \DA{mentioned it as an interactive appl}

%\AL{We should have a teaser figure summarizing what we do in the paper. Dataset is a part, but in the current version, method is another important part to claim. }

\iffalse
While visual language models (VLMs) are increasingly demonstrating strong abilities in image captioning and visual question answering (VQA) \RRM{cite vqa}, these models are increasingly faced with more difficult tasks that require reasoning about the spatial relationships within an image \RRM{cite}. 
\fi
Spatial reasoning is essential for humans to interact with the world, such as determining if there is enough room on a desk for a backpack; if there is enough space to navigate through a room without hitting any obstacles; or if an object is placed sufficiently high enough to be inaccessible to a child. 
%Humans can intuitively estimate the sizes of and distances between specific objects in an environment from a distance and \textit{without} using external tools by combining visual cues with respect to available objects in the scenes with commonsense reasoning \citep{gunzelmann2006qualitative}. 
Spatial reasoning skills are also essential for visual AI agents in interactive applications such as visual question answering, augmented reality, robotics, or visual assistants~\cite{rocamonde2024visionlanguage,du2023visionlanguage}.
% \DA{I would like to mention here sth like While estiamting distances from monocular images is ill-posed, Humans .... . We want to let clear we known estimating distances from 2d images is ill-pose  yet ...... .}\AL{I moved this to the next paragraph. Please review.}
%Thus, direct spatial reasoning is a requirement for visual AI agents in interactive applications such as visual question answering (VQA), augmented reality, robotics, or visual assistants~\cite{rocamonde2024visionlanguage,du2023visionlanguage}.

Despite this need, the ability for an off-the-shelf vision language model (VLM) to perform fine-grained spatial reasoning remains under-explored. 
Current benchmarks primarily assess whether these models understand qualitative concepts like ``left'' versus ``right'' or ``near'' versus ``far'' from monocular images; formulating these tasks as question-answering~\cite{Johnson2016CLEVRAD, Krishna2016VisualGC, Liu2022VisualSR, kamath2023whatsup}. However, recent studies have revealed that state-of-the-art VLMs struggle with quantitative spatial tasks. 
While measuring sizes or distances from monocular images is ill-posed~\cite{Saxena2005LearningDF}, humans are surprisingly adept at making such estimations by relying on contextual clues~\cite{sedgwick1986space}, suggesting the potential improvements in VLMs.

To benchmark quantitative spatial reasoning, Spatial VQA~\cite{spatial_vlm} and SpatialRGPT-bench~\cite{spatial_rgpt} are proposed. Yet, both benchmarks either heavily relies on metric depth estimation or on the coarse label representations, \eg, 3D cuboids, potentially leading to inaccurate quantitative labels.
On the other hand, to improve quantitative spatial reasoning in VLMs, Spatial VLM and SpatialRGPT demonstrate that \emph{fine-tuning} VLMs with additional spatial VQA data and architectural modifications can enhance their performances significantly. However, this requires direct access to the model, as well as a significant amount of data and computational resources.

\begin{figure*}[t]
\centering
\includegraphics[width=\textwidth]{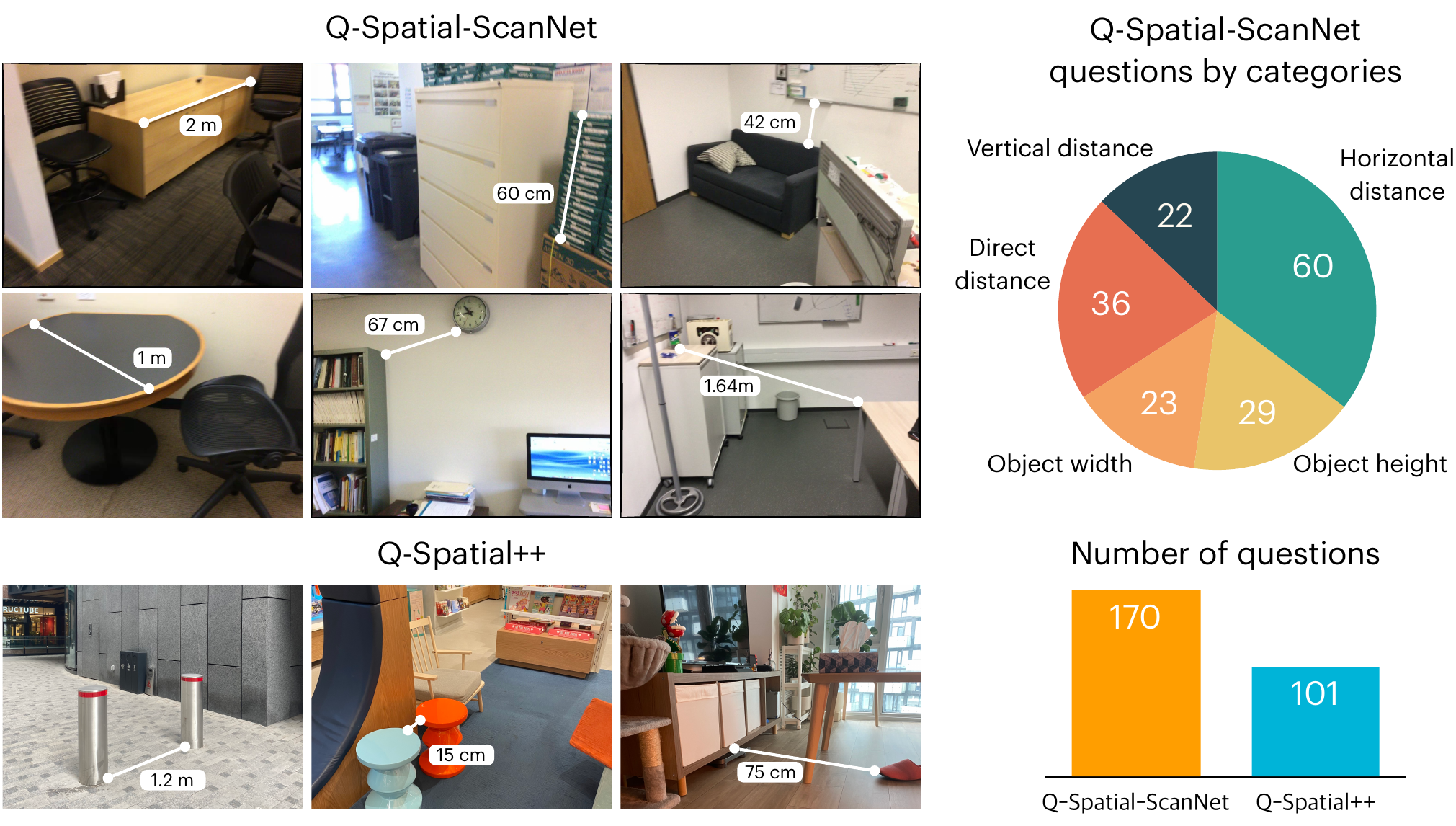}
\caption{We introduce a human expert-annotated benchmark dedicated to quantitative spatial reasoning: \textbf{\benchmark}. The benchmark consists of two splits: \scannetqas and \inhouseqas. The left panel shows the examples from the two splits. \scannetqas is repurposed from a subset of images and RGB-D scans in ScanNet~\cite{Dai2017ScanNetR3} and the questions are categorized into five categories (top-right). To provide a more robust evaluation in quantitative spatial reasoning, we captured an additional set of images and provide accurately-annotated quantitative spatial questions for \inhouseqas.
%\DA{let's be more specific about ours here, e.g. it is newly annotated and highly accurate using measure tape} %\DA{last sentence}\AL{need to change the numbers in the bottom right figure.
}\label{fig:bench_stats}
\end{figure*}

In this paper, we introduce a new question answering corpora, \benchmark, specifically designed to evaluate quantitative spatial reasoning in VLMs with high-precision. We identify the strengths and weakness across models by systematically analyzing the performance of SoTA VLMs.
In particular, our analysis shows that VLMs become more effective when a reasoning path involving reference objects emerges in the responses. 
Inspired by this, we propose a prompt-based technique, \OurMethod, to guide VLMs to effectively solve quantitative spatial tasks.
In contrast to prior works~\cite{spatial_vlm,spatial_rgpt}, \OurMethod bypasses the need for direct access to the model, additional task-specific data, architectural modifications, or fine-tuning.

We demonstrate that by solely using \OurMethod, \gemini, \geminiflash, and \gptv improve their success rates by over 47, 22, and 30, respectively, in \benchmark \textit{without} the need for more data, model architectural modifications, or fine-tuning.
To put these numbers in perspective, we note that the pioneering work, SpatialVLM \cite{spatial_vlm}, was fine-tuned on a massive dataset with 2 billion direct spatial reasoning QA pairs to achieve a relative improvement of less than 4 points in the same metric. % \DA{please check this claim, table 2 second row in spatialvlm palm2e vs theirs}. 
While their evaluation benchmark is not publicly available for a direct comparison, we could still use this number as a proxy for reference.
We emphasize our improvement  is achieved purely with simple instructions using natural language. There is no external training data, model
architectural modifications, or fine-tuning.
Most importantly, we believe our approach of analyzing emergent reasoning paths in high-performing models opens new horizons for transferring capabilities to less performant ones in a cost-effective manner.
\section{Related Works}

\subsection{Spatial Reasoning Benchmarks}
%\AL{Benchmark papers to cite: ~\cite{Cai2024SpatialBotPS}}

Many vision-language benchmarks evaluate spatial relationships~\cite{Agrawal2015VQAVQ,Johnson2016CLEVRAD,Krishna2016VisualGC,Suhr2018ACF,Yang2019SpatialSenseAA,belz-etal-2018-spatialvoc2k,Goyal2020Rel3DAM,OpenEQA2023}. 
%, such as VQA~\cite{Agrawal2015VQAVQ,Goyal2016MakingTV}, CLEVR~\cite{Johnson2016CLEVRAD}, VisualGenome~\cite{Krishna2016VisualGC}, NLVR2~\cite{Suhr2018ACF}, OpenEQA~\cite{OpenEQA2023} \etc. 
However, spatial reasoning is entangled with other challenges in these datasets. With the promise of deploying vision-language models in the physical world~\cite{rt12022arxiv,rt22023arxiv}, there are more benchmarks designed to evaluate spatial reasoning alone. Most of them are tasked to evaluate qualitative spatial reasoning where models are tasked to recognize spatial relationships between objects in 3D space, such as ``left'' versus ``right'' or ``behind'' versus ``in front of''. VSR~\cite{Liu2022VisualSR} extend this direction by increasing linguistic complexities. What’sUp~\cite{kamath2023whatsup} is designed to remove spatial biases in the real-world.

Quantitative spatial reasoning, where models are tasked to estimate quantitative spatial information such as sizes and distances, is relatively underexplored. Spatial VQA~\cite{spatial_vlm} contains both qualitative and quantitative spatial questions. However, it is publicly unavailable. SpatialRGPT-Bench~\cite{spatial_rgpt} is an open-source benchmark and follows the similar data-generation pipeline in Spatial VQA. Both benchmark are first automatically-generated from templates and followed by human verification.
Our work takes a step further by providing a human-annotated benchmark dedicated to quantitative spatial reasoning, providing high-precision measurements. Moreover, we capture our own images to minimize the risk of test data leakage~\cite{elangovan-etal-2021-memorization}.

\subsection{Spatial Reasoning in VLMs}

Many prior works improve spatial reasoning by leveraging depth information~\cite{Fichtl2014LearningSR,Rosman2011LearningSR} or sequential RGB images~\cite{Kase2020TransferableTE,Yuan2021SORNetSO,Migimatsu2021GroundingPT,OpenEQA2023}. Our work focuses on spatial reasoning from a single monocular image. 
SpatialVLM~\cite{spatial_vlm} and SpatialRGPT~\cite{spatial_rgpt} are most aligned to our work. 
SpatialVLM propose an automatic data engines to generate training data for VLM fine-tuning. 
SpatialRGPT introduces a plugin module for relative depth input to further boost spatial reasoning in VLMs.
Unlike prior works, we propose a prompting technique that consistently improves quantitative spatial reasoning in VLMs, bypassing the need for direct access to the model, additional task-specific data, architectural modifications, or fine-tuning.

\begin{table}[t]
\centering
\resizebox{\columnwidth}{!}{%
\begin{tabular}{@{}lrrr@{}}
\toprule
                                      & \multicolumn{1}{c}{\begin{tabular}[c]{@{}c@{}}\# Quantitative\\ spatial questions\end{tabular}} & \multicolumn{1}{c}{\begin{tabular}[c]{@{}c@{}}Human\\ annotated\end{tabular}} & \multicolumn{1}{c}{\begin{tabular}[c]{@{}c@{}}Publicly\\ available\end{tabular}} \\ \cmidrule(l){2-4} 
Spatial VQA~\cite{spatial_vlm}        & 215                                                                                             &                                                                               &                                                                                  \\
SpatialRGPT-Bench~\cite{spatial_rgpt} & 749                                                                                             &                                                                               & \checkmark                                                                       \\
\benchmark \small{(ours)}             & 271                                                                                             & \checkmark                                                                    & \checkmark                                                                       \\ \bottomrule
\end{tabular}%
}
\caption{\textbf{Comparison of quantitative spatial reasoning benchmark.} \benchmark is a human expert-annotated benchmark, specifically designed for quantitative spatial questions.}
\label{tab:benchmark_comp}
\end{table}

%\section{Benchmark: \benchmark}\label{sec:benchmark}
\section{Quantitative Spatial Reasoning}

\subsection{Preliminaries}

%\RRM{I am not sure what this preliminaries gives us that warrants a separate section. Is it possible to move to a subsection of \S4? Up to you, if you feel strongly then leave it as is}
%\AL{Merge preliminaries here.}

We explore quantitative spatial reasoning where a VLM is tasked to recognize quantitative spatial information of physical objects such as distances or sizes from a 2D image. In particular, we consider \textit{direct} quantitative spatial reasoning, where a VLM predicts the quantitative spatial information \textit{without} accessing any external tools or large models. 
%In other words, we are interested in the potential depth and 3D structures infused in the weights in VLMs.

%This work focuses on quantitative spatial reasoning, where the objective is to recognize quantitative information of physical objects such as distances or size differences from a 2D image. While this is not a novel task and it usually considered as a part of broader applications, \eg, SLAM~\cite{DurrantWhyte2006SimultaneousLA} and depth estimation~\cite{fu2018deep}, we are interested in the potential depth and 3D structures infused in the weights in large vision-language models~\cite{spatial_vlm}. 
%\DA{sentence above might feel vague}\AL{please review} \RRM{I dont like that last sentence (vague). I think it is fine to remove it and merge the above and below paragraphs.}
%In other words, we are interested in performing \textit{direct} quantitative spatial reasoning via VLMs.

Following prior work~\cite{spatial_vlm}, we can formulate quantitative spatial reasoning into a question answering task. For an image $I$ and a question querying spatial quantities in the image $Q$, the model needs to output the estimated quantities in text format 
Specifically, given an image and a question querying spatial quantities in the image, the model needs to directly output the estimated quantities in text format $A$. 
%\DA{Lets define this with notation I and Q, A as spatial vlm did in 3.3}\AL{please review}

\subsection{Benchmark: \benchmark}\label{sec:benchmark}
%In this section, we form the quantitative spatial reasoning question-answering problem~\cite{spatial_vlm} and introduce a human expert-annotated benchmark dedicated to quantitative spatial reasoning: \benchmark.
%In this benchmark, a VLM answers questions that require measuring distances in the real world through a 2D image. Specifically, we focus on \textit{direct} quantitative spatial reasoning, where a VLM takes an image and a question involving a quantitative spatial task and provides an answer as a text string without using external tools or interacting with other large models~\cite{spatial_vlm}. 
%\AL{A little bit out of place?}\DA{I agree, i feel we need to introduce first thet task, I think we need a Preliminary/Background section before proposing the benchmark, the nuances between direct , qualitative and quantitative should be clearly introduced. it is fine if its short }
%\RRM{how much can we put into \S2?}
%\subsection{\benchmark}

Existing benchmarks in quantitative spatial reasoning are either not publicly available~\cite{spatial_vlm} or automatically-generated~\cite{spatial_rgpt}. We introduce a human expert-annotated benchmark dedicated to quantitative spatial reasoning: \benchmark, highlighting its high-quality and high-precision question-answer pairs. Additionally, the benchmark provides a split with freshly captured images which are unseen to the commercial VLMs. Table~\ref{tab:benchmark_comp} compares \benchmark with existing benchmarks.
Though being manually annotated, \benchmark is as large as Spatial VQA and will be made publicly available.

\benchmark consists of two splits: \scannetqas and \inhouseqas. 
Each split contains a series of image and annotated question tuples. 
We consider five different categories of spatial questions estimating
(i) the width of an object; (ii) the height of an object; (iii) the horizontal distance between two objects on the same plane; (iv) the vertical distance between two objects vertically separated from each other; and (v) the direct distance between two objects in the scene. 
This categorization allows us to dissect the strengths and weaknesses of VLMs.
Fig.~\ref{fig:bench_stats} shows examples of these questions and overall statistics of \benchmark.
% \DA{This is only true for the scannet split right?, it reads as if the two splits have the same five categories. }
% \RRM{I think it's okay to have all 5 categories introduced here and then specify that in \inhouseqas, we only focus on distances between objects. the categories are a general thing and perhaps future work will explore the other categories in more detail. otherwise, we casually refer to categories before explicitly really defining them anywhere and i think it is good to have them defined upfront, even if we only focus on some of them.}
% \DA{OK}

\textbf{\scannetqas} is repurposed from the high-quality RGB-D scans of real-world environments in ScanNet~\cite{Dai2017ScanNetR3}.
\scannetqas consists of 99 images and 170 human expert-annotated questions. For each question, the annotators are instructed to navigate in the point clouds provided in the dataset to measure the metric distances. 
%We categorize the questions into five classes including measuring width and height of an object and a horizontal, vertical, or direct distance between two objects. For example, ``What is the width of object A?'', ``What is the minimum distance between object A and B?'' This allows us to dissect VLMs' strengths and weakness in quantitative spatial reasoning in Sec.~\ref{sec:analysis}.

The annotation pipeline consists of three stages: image selection, distance annotation, and QA generation. First, we selected a subset of images from ScanNet as our candidate images. We exclude images that are blurry or with extremely simple layout, \eg, a corner of a room without any objects.
%\DA{Based on what the selection was done?}
Next, with the provided reconstructed point clouds in ScanNet, we perform voxel downsampling to the point clouds and visualize them with Plotly~\cite{plotly}. This allows human experts to navigate through the point clouds and annotate the end points of the distances.
%we ask human experts to navigate the point clouds and annotate end points of the distances between various objects and object pairs. 
%\DA{with what tools, how?}
The experts then describe the annotated distances in natural language.
As compared to the prior work~\cite{spatial_rgpt} that compute the distances from 3D cuboids, allowing human experts to explicitly specify the end points of a distance leads to higher-precision distance measurements
%By allowing human experts to explicitly annotate the end points to specify the distances, we 
%we are allowed to measure the distances such as the minimal/maximal distance between object A and B.
%This fine-grained distance representations ensure the high-precision in the benchmark.
% \DA{why, it was not clear to me why is more precise from the description given?}\AL{please review}
% 
With these annotations, we construct a triplet of an image, a distance quantity, and a description of the distance. Finally, we leverage Llama 3-70b~\cite{llama3modelcard} to convert the triplets to the question-answering pairs and manually verify the sanity of each pair.
%\DA{The output of the model was then verified and in some cases rectified by a human  to ensure high-quality.}
% \DA{maybe we should emphasize the role of llama3 here, it should be clear the measurements are done by humans. }\AL{Please review.}
% We consider five different categories of spatial questions in this split:
% (i) estimating the width of an object; (ii) estimating the height of an object; (iii) estimating the horizontal distance between two objects on the same plane; (iv) estimating the vertical distance between two objects vertically separated from each other; and (v) estimating the minimum direct distance between two objects in the scene. 
% This diverse set of question allows us to dissect the strengths and weaknesses of VLMs.
% \DA{Added the different categories of questions here since it is only on this split we did that}

\textbf{\inhouseqas} is designed to provide a robust evaluation of VLMs for quantitative spatial reasoning. 
Because ScanNet is a pre-existing publicly available dataset, it is uncertain whether existing commercial VLMs were trained on these images. 
To mitigate this uncertainty, \inhouseqas uses 101 diverse images captured with an iPhone in various environments, including indoor, outdoor, daytime, and nighttime settings. %\DA{What are the proportions, maybe put in percetange when mentioning in the line above}\AL{TODO}
Each image was captured specifically for spatial question answering, and has several objects within. While capturing these photos, we physically measured the distances between objects in the real-world. %with high-precision tool 
%\DA{what tools, digital or in the real-world? why are they high precision?}. 
For this split, we focus only on horizontal distances between objects, which we observed more challenging for VLMs (see Section \ref{sec:analysis}).
Overall, there are 101 annotated questions in this split.
%\DA{Let's be specific what type of distance, horizontal distances?}
%We annotated 71 questions focused on measuring distances between objects, a task identified as challenging for VLMs in Section \ref{sec:analysis}. 
%The answers to these questions were measured using physical tools to ensure high precision.

\subsection{Evaluation metrics}
In \benchmark, a VLM must provide the answer in text format to facilitate regex extraction with a fixed script. We adopt a consistent output format for every VLM. We measure the performance in success rate by thresholding the maximum ratio between an estimation and a ground truth value. Given a distance estimation $\hat{d}$ and a ground truth value $d^*$, the maximum ratios between the estimation and ground truth can be expressed as $\delta = \max(\frac{\hat{d}}{d^*},\frac{d^*}{\hat{d}})$. Following existing works~\cite{spatial_vlm,spatial_rgpt}, we set the thresholds to $1.25$ and $2$ and denote the evaluation metrics as $\delta_{\leq 1.25}$ and $\delta_{\leq 2}$, respectively. In other words, an accurate answer should be within $[0.8\times, 1.25\times]$ or $[0.5\times, 2\times]$ the ground truth value, respectively. We use $\delta_{\leq 2}$ as the main evaluation metrics in this paper and report $\delta_{\leq 1.25}$ in the Appendix.

\begin{table}[t]
\centering
\resizebox{\columnwidth}{!}{%
% \begin{tabular}{@{}lrrrr@{}}
% \toprule
%             & \llava & \gemini & \gptv & \gpto \\ \cmidrule(l){2-5} 
% \scannetqas & 60.6   & 1.6$^*$ & 25.3  & 64.13 \\
% \inhouseqas & 38     & 5$^*$   & 24.42 & 62.91 \\ \bottomrule
\begin{tabular}{@{}lrrrr@{}}
\toprule
            & Gemini 1.5 Pro$^*$& Gemini 1.5 Flash & \gptv & \gpto \\ \cmidrule(l){2-5} 
\scannetqas & 0.59& 20 &28.24& \textbf{69.41}\\
\inhouseqas & 0.99& 26.73&18.81& \textbf{61.06}\\ 
Overall & 0.79& 23.36&23.52& \textbf{65.23}\\ \bottomrule
\end{tabular}%
}
\caption{\textbf{\gpto outperforms other commercial VLMs in quantitative spatial reasoning.} We evaluates the success rate $\delta_{\leq 2}$ on each split of \benchmark.
$^*$\gemini consistently refuses to provide the measurements.
%\RRM{Should we have an overall row to show aggregate perf?}\AL{Sure. Do we aggregate in QA instance level or in dataset level? I think it should be QA instance level}\RRM{IMO doesn't matter as long as we specify how we did it somewhere}
}
\label{tab:preliminary_analysis}
\end{table}
\begin{table}[t]
\centering
\resizebox{\columnwidth}{!}{%
\begin{tabular}{@{}lrrrr@{}}
\toprule
%                     & \llava & \gemini & \gptv & \gpto \\ \cmidrule(l){2-5} 
% Object width        & 78.3   & 0       & 43.4  & 81.1  \\
% Object height       & 62.1   & 1.5     & 51.5  & 90.9  \\
% Horizontal distance & 53.3   & 2.7     & 15    & 42.7  \\
% Vertical distance   & 41.4   & 0       & 35.6  & 67.83 \\
% Direct distance     & 75     & 1.86    & 41.66 & 69.43 \\ \bottomrule
                    & Gemini 1.5 Pro$^*$ & Gemini 1.5 Flash  & \gptv & \gpto \\ \cmidrule(l){2-5} 
Object width        & 0.87& 43.48& 44.93& \textbf{81.16}\\
Object height       & 0& 31.82& 30.3& \textbf{93.94}\\
Horizontal distance & 0.33& 13.33& 10& \textbf{49.44}\\
Vertical distance   & 0.69& 20.69& 33.33& \textbf{71.26}\\
Direct distance     & 1.11& 8.33& 42.59& \textbf{78.7}\\ \bottomrule

\end{tabular}%
}
\caption{\textbf{Success rate $\delta_{\leq 2}$ breakdown by question categories.} Among five question categories, measuring the distance between two objects is more challenging for SoTA VLMs, particularly when measuring horizontal distances. $^*$\gemini consistently refuses to provide the measurements. %\AL{Can we put $^*$ only on one number?} \RRM{Just * the Gemini column header}
}
\label{tab:vlm_strength_weakness}
\end{table}

\begin{figure*}[t]
\includegraphics[width=\textwidth]{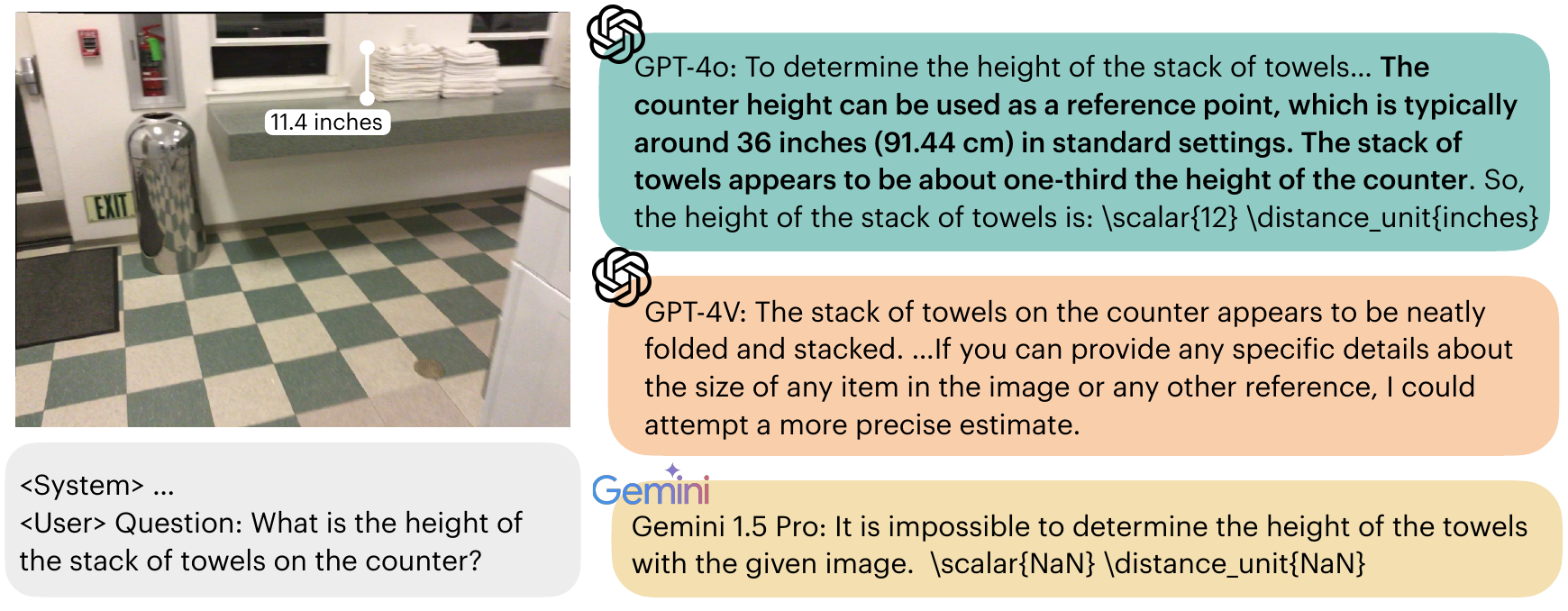}
\centering
\caption{\textbf{Qualitative analysis of different VLMs responses.} As compared to \gptv and \gemini, we observe that for many correct instances, a reasoning path  using a reference object naturally emerges  in \gpto responses.
% We also make the surprising observation that the performance of the top-performing VLM increases when a reasoning path using a reference object emerges naturally in the response.
% 
%\RRM{The image states ground truth at 29cm but the GPT-4o response is in inches. Can we change the image to inches (or convert the GPT-4o response to cm?) Otherwise, a reader not paying attention will wonder that GPT-4o's response is poor!}
A \textit{reference object} is another object in the scene for whose spatial dimensions can be easily inferred via commonsense reasoning. 
Exploiting this observation, we design a zero-shot prompting technique that encourages both underperforming  VLMs \textit{and} \gpto itself to reason this way. 
}\label{fig:example_outputs}
\end{figure*}

\section{Analysis}\label{sec:analysis}

We first test four commercial VLMs including \gemini, \geminiflash~\cite{geminiteam2024gemini}, \gptv~\cite{openai2024gpt4}, and \gpto, to identify which question categories are more challenging or easier for VLMs in Sec.~\ref{sec:strength_and_weakness}.  
%We then statistically analyze why GPT-4o outperforms the other models on all tasks.
We then statistically analyze the potential factors leading to the strong \gpto performances on all tasks in Sec.~\ref{sec:why_gpt4o_is_good}.

\subsection{Strengths and Weaknesses}\label{sec:strength_and_weakness}
% \noindent \textbf{Main results.} 

Table~\ref{tab:preliminary_analysis} evaluates the success rate $\delta_{\leq 2}$ in \benchmark.
\gpto outperforms all other models on both splits with success rates of 69.41  and 62.44, respectively. The performance gap between \gpto and the second-best model, \gptv, is significant by a margin of over 41 points. 
%We also notice that all models perform slightly better in the \scannetqas vs. our newly collected images.
%\DA{I am assuming LLava does not exist here}

% \noindent \textbf{Strengths and Weaknesses.} 
To further understand the strengths and weakness within each VLM, we break down \scannetqas by question categories and demonstrate the success rate $\delta_{\leq 2}$ in Fig.~\ref{tab:vlm_strength_weakness}. In particular, ``Horizontal distance'' questions are challenging for most VLMs, and usually achieve the lowest success rates. \gpto leads in all question types, showing the highest success rates, especially in ``Object width'' and ``Object height'' with 81.16 and 93.94 points, respectively. 
%\AL{We dont have blind LLM baseline here. Otherwise, we can further say that it's probably due to the fact that these types of question can be answered through common sense.}
The overall challenge of estimating vertical and horizontal distances is evident, as these categories have the lowest success rates across models, particularly for \gemini, which we observe consistently refuses to provide  measurements. We consider the responses with unspecified quantities as incorrect answers.
%\RRM{How do we evaluate Gemini when it doesn't provide measurements? Do we treat them as automatic error? Do we omit these from calculations? Please specify}
Notably, the gap between GPT-4o and the second-best model on ``Horizontal distance'' question is over 35 points.

%\DA{I'd remove Llava from here too}
%\DA{If we keep LLava,the section above requires discussion into why Llava is good. It should also mention we did not observe anything special about it.... }
%\RRM{We should have a version of Table 3 that doesn't have reasonign prompts and use this to demonstrate that GPT-4o is good.}\AL{added}

\subsection{Understanding the dominance of GPT-4o}\label{sec:why_gpt4o_is_good}

A qualitative analysis of the output text generated by GPT-4o reveals that for many correct instances, the output text suggests a reasoning path using one or more reference objects to guide the spatial estimation of the target object in each scene. A reference object is another object in the scene for whose spatial dimensions can be easily inferred via commonsense reasoning. 
As shown in Fig.~\ref{fig:example_outputs}, when measuring the height of the stack of towels, the counter near it can act as a reference object. While the height of the stack of towels can have variable height, the counter appears to have a standard height (36 inches suggested by \gpto). 
%\DA{Can we put in brackets e.g.(the standard height of a counter is XXX?}\AL{added}

%For example, when measuring the width of a sofa (target object) in a room, the door beside the sofa can act as a reference object (see Figure \ref{fig:bench_stats}). While a sofa can have some variable width, the door appears to have a standard width and height. 
%From a reasoning perspective, the door introduces additional facts about the scene, and analyzing the relative position of the sofa with respect to the door can lead to a more accurate estimate than by na\"ively estimating the size of the sofa in isolation. 
From a reasoning perspective, the height of the counter introduces additional facts about the scene, and analyzing the relative position of the stack of towels with respect to the counter can lead to a more accurate estimate than by na\"ively estimating the size of the stack of towels in isolation.

\begin{table}[ht]
\centering
\resizebox{0.85\columnwidth}{!}{%
\begin{tabular}{@{}lrr@{}}
\toprule
               & Correct ($\delta_{\leq 2}$) & Incorrect ($\delta_{> 2}$) \\ \cmidrule(l){2-3} 
w/ references  & 45                   & 9                      \\
w/o references & 114                  & 63                     \\ \bottomrule
\end{tabular}%
}
\caption{\textbf{Contingency table} of whether \gpto's responses use any reference objects as guidance and the success rate of the responses. GPT-4o used a reference object in 23.4\% of instances. However, for the 54 questions where a reference object was used, the success rate $\delta_{\leq 2}$ is 83\% (45/54), compared to 64\% (114/177) when no reference object was used.}
\label{tab:gpt4o_analysis_contingency_table}
\end{table}

To quantify the above observation, we first count the number of instances where the \gpto output involves reasoning via a reference object to answer the question. To do this, we feed the sampled \gpto responses to a separate \gpto instance which we prompt to determine if a model response uses any reference object as guidance (see Appendix~\ref{supp:full_text_prompt} for details). Table \ref{tab:gpt4o_analysis_contingency_table} gives the contingency table of the number of instances where at least one reference object was used versus whether the maximum ratio $\delta$ is less 2.
% 200\% (i.e., $\delta < 2$). 
Note that \gpto used a reference object for 23.4\% of instances. However, of the 54 questions where a reference object was used, the success rate $\delta_{\leq 2}$ is 83\%, versus the success rate $\delta_{\leq 2}$ of only 64\% over instances when a reference object was not used. This motivates the following hypothesis:

\begin{center}
    \emph{Reasoning via a reference object directly improves overall accuracy on quantitative spatial questions.} 
\end{center}

To test our hypothesis, we fit a logistic regression model on the \gpto responses to %instances and 
whether the estimate is accurate. For any response, let $X_r \in \{0,1\}$ indicate whether the response used any reference object. 
To isolate the effect of using any reference object, we control for two task-related factors. 
First, it may be that one of the two dataset splits is substantially easier than the other, thereby leading to more accurate estimates; here, let $X_d \in \{0,1\}$ indicate whether the question belongs to the \inhouseqas split or \scannetqas split. 
Second, tasks where the ground truth distance measured is large inherently have a larger absolute range to achieve $\delta_{\leq 2}$. To ensure that \gpto is not just better at answering questions involving larger distances, let $X_g$ be the ground truth distance (in cm) for each question. Thus, our logistic regression model fits:
\begin{equation}\nonumber
    p(\delta_{\leq 2}) \sim \beta_0 + \beta_r X_r + \beta_d X_d + \beta_g X_g 
\end{equation}
where $\beta_0, \beta_r, \beta_g, \beta_d$ are regression coefficients.

% \usepackage{booktabs}
% \usepackage{graphicx}
\begin{table}[t]
\centering
\resizebox{\columnwidth}{!}{%
\begin{tabular}{@{}lcc@{}}
\toprule
                           & Coefficient $\beta$ & SE \\ \cmidrule(l){2-3} 
Use reference object $(X_r)$       & 1.0179$^*$     & 0.406 \\ 
From \inhouseqas $(X_d)$    & 0.0276      & 0.318 \\
Ground truth distance (cm) $(X_g)$ & 0.0017      & 0.002 \\
\bottomrule
\end{tabular}%
}
%\begin{tabular}{@{}lcc@{}}
%\toprule
%                           & Coefficient & p-value \\ \midrule
%Ground truth distance (cm) & 0.0017      & 0.452   \\
%Data Source = ours         & 0.0276      & 0.931   \\
%Use references = True      & 1.0179*     & 0.012   \\ \bottomrule
%\end{tabular}%
\caption{\textbf{Logistic regression to analyze the effectiveness of GPT-4o.} $^*$ denotes a $p$-value $< 0.05$. Using a reference object in reasoning increases the likelihood of generating a response with relative error $\delta$ less than 2, statistically significantly. 
%\RRM{It is a bit confusing when you use $\delta \leq 2$ and $ \delta_{\leq2}$. Can we stick to one notation?}
}
\label{tab:gpt4o_analysis}
\end{table}

Table \ref{tab:gpt4o_analysis} summarizes the results. We first find that when \gpto uses a reference object in the responses, the odds of an accurate estimate increase by a factor of $e^{1.0179} \approx 2.7$ ($p$-value $<0.05$). 
Furthermore, there is no statistically significant difference in terms of \gpto performance on \inhouseqas or \scannetqas, \ie, the model performs roughly equally well on both splits. 
Finally, the accuracy of the distance estimate does not statistically significantly depend on the actual ground truth distance. 
That is, even if the question involves measuring a large distance, this does not change the likelihood of generating an accurate response, after factoring for whether a reference object is used. \textit{We conclude by verifying our hypothesis that reference objects can improve the overall reasoning of the model.}

\section{\OurMethod: Eliciting Reasoning Paths with Reference Objects}\label{sec:spatial_prompt}

%\RRM{This section feels oddly short. How do we feel about moving this to a subsection in \S4? or a subsection in \S6? At the bare minimum I don't think \S5.1 needs to be a separate subsection.}

Inspired by the analysis in Sec.~\ref{sec:why_gpt4o_is_good}, we propose a prompting technique, \OurMethod, for quantitative spatial reasoning that explicitly encourages VLMs to identify reference objects within the images.
Orthogonal to prior work~\cite{spatial_vlm,spatial_rgpt}, our prompting techniques involve no additional data generation, fine-tuning or architectural modifications.

%Inspired by the reasoning paths observed in \gpto in Sec.~\ref{sec:why_gpt4o_is_good}, we ``distill'' the reasoning paths to models, we introduce a method for "distilling" those using natural language instructions. Unlike traditional distillation, which typically involves transferring knowledge from a larger model to a smaller one through supervision, our approach leverages these strategies to guide underperforming models via zero-shot prompt-based instructions.

\subsection{Prompt Design}

% We propose 
\OurMethod is specifically designed to trigger a reasoning path  that uses  reference objects in the image as visual cues. 
% improves the quantitative spatial reasoning capabilities in VLMs by encourage them to reason by identifying reference objects within the images. 
% Due to the specialized domain in quantitative spatial reasoning, designing a simple prompt that works for every VLM is challenging. \DA{Consider removing the previous sentence.} 
We propose two variants of text prompts: \OurMethodSimple and \OurMethodComplex as shown in Fig.~\ref{fig:main_prompt}. The former is inspired by zero-shot CoT~\cite{kojima2022large}. We specifically design \OurMethodSimple to keep it as compact and easy to remember as possible. 
In contrast, \OurMethodComplex provides a more detailed breakdown of the steps to solve the problem. We found that some VLMs perform better when given explicit step-by-step instructions. For instance, \gemini follows the detailed instructions more effectively, whereas \gptv's performance is less dependent on the complexity of the prompt as long as it is encouraged to find a reference object. %\DA{Check this is correct} 

% Thus, we design \OurMethodComplex as an alternative.

% For VLMs that are more resilient to prompt changes\DA{that are better at following instructions?} in quantitative spatial questions, we design \OurMethodComplex as a stronger text prompt to guide the reasoning.

\begin{figure}[t]
\centering   
\small{\OurMethodSimple}
\begin{tcolorbox}[left=2pt, right=2pt, top=1pt, bottom=1pt, boxrule=1pt]
\tiny{
\begin{Verbatim}[breaklines=true, commandchars=\\\{\}]
User: Question: \textcolor{red}{[SPATIAL QUESTION]}
Let's think step by step and start by finding good reference objects or object parts in the image.
Assistant: {\textcolor{red}{[OUTPUT]}}
\end{Verbatim}
}
\end{tcolorbox}
\small{\OurMethodComplex}
\begin{tcolorbox}[left=2pt, right=2pt, top=1pt, bottom=1pt, boxrule=1pt]
\tiny{
\begin{Verbatim}[breaklines=true, commandchars=\\\{\}]
User: Question: \textcolor{red}{[SPATIAL QUESTION]}
Use the following 4 steps sequentially to answer the question:
Step 1 **Analyze the question**

Step 2 **Identify up to 10 reference scales in the image, ranging from large to small sizes, and list them in the specified format**
[details omitted]

Step 3 **Propose a robust step-by-step plan to answer the question by using the reference scales in Step 2**
[details omitted]

Step 4 **Focus on the image and follow the plan in Step 3 to answer the question**
Assistant: {\textcolor{red}{[OUTPUT]}}
\end{Verbatim}
}
\end{tcolorbox}
\caption{We propose \OurMethod, a specialized text prompt designed to improve quantitative spatial reasoning capabilities in VLMs by encourage them to reason by identifying reference objects within the images.}\label{fig:main_prompt}
\end{figure}

\begin{table*}[t]
\centering
\resizebox{\textwidth}{!}{%
\begin{tabular}{@{}lrrrrrrrr@{}}
\toprule
 &
  \multicolumn{4}{c}{\scannetqas} &
  \multicolumn{4}{c}{\inhouseqas} \\
 &
  \multicolumn{1}{c}{Gemini 1.5 Pro} &
  \multicolumn{1}{c}{Gemini 1.5 Flash} &
  \multicolumn{1}{c}{GPT-4V} &
  \multicolumn{1}{c}{GPT-4o} &
  \multicolumn{1}{c}{Gemini 1.5 Pro} &
  \multicolumn{1}{c}{Gemini 1.5 Flash} &
  \multicolumn{1}{c}{GPT-4V} &
  \multicolumn{1}{c}{GPT-4o} \\ \cmidrule(l){2-9} 
Standard prompt &
  0.59  &
  20  &
  28.24  &
  69.41  &
  0.99&
  26.73&
  18.81&
  61.06\\
Zero-shot CoT &
  2.94  &
  18.53  &
  \textbf{56.47}&
  70.2  &
  4.16&
  26.11&
  46.53&
  59.74\\
\OurMethod &
  \textbf{53.65}&
  \textbf{52.71}&
  54.9  &
  \textbf{71.96}&
  \textbf{43.17}&
  \textbf{38.42}&
  \textbf{53.47}&
  \textbf{62.71}\\ \bottomrule
\end{tabular}%
}
\caption{\textbf{Success rate $\delta_{\leq 2}$ of different VLMs and prompting techniques.} The proposed prompt \OurMethod consistently leads to higher success rates across different VLMs. 
We bold font the best numbers across different prompting techniques and highlight their performances as compared to the performances of the standard prompt.}
\label{tab:main_results}
\end{table*}

% \usepackage{booktabs}
% \usepackage{graphicx}
\iffalse
\begin{table*}[t]
\centering
\resizebox{\textwidth}{!}{%
\begin{tabular}{@{}lrrrrrr@{}}
\toprule
                & \multicolumn{3}{c}{\scannetqas}                                                     & \multicolumn{3}{c}{\inhouseqas}                                                     \\
                & \multicolumn{1}{c}{\gemini} & \multicolumn{1}{c}{\gptv} & \multicolumn{1}{c}{\gpto} & \multicolumn{1}{c}{\gemini} & \multicolumn{1}{c}{\gptv} & \multicolumn{1}{c}{\gpto} \\ \cmidrule(l){2-7} 
Standard prompt & 6.4                         & 41.1                      & 28.9                      & 4.2                         & 24.6                      & 11.4                      \\
Zero-shot CoT   & 7                        & 65.6                      & 46.2                      & 0                           & 61.2                      & 39                        \\
\OurMethod \small{(ours)}      & \textbf{98.8}                        & \textbf{76.8}                      & \textbf{90}                        & \textbf{97.1 }                       & \textbf{76.4 }                     & \textbf{81.5}                      \\ \bottomrule
\end{tabular}%
}
\caption{\textbf{Frequency of whether the responses involve using reference objects of different VLMs and prompting techniques.} The proposed prompt \OurMethod consistently lead to higher chances to have reference objects involved in the responses. }
\label{tab:compare_reference_frequency}
\end{table*}
\fi

\subsection{Experiments}

In this section, we evaluate the effectiveness of the proposed text prompting techniques in success rate $\delta_{\leq 2}$ in Sec.~\ref{sec:main_results}. Since our initial motivation is to encourage VLMs to use reference objects within the images more frequently, we report the frequency of whether the VLMs use reference objects in the responses in Sec.~\ref{sec:frequency}. Lastly, we discuss the attempt to apply our prompting techniques to open-source VLMs in Sec.~\ref{sec:open_source}.

To gauge the effectiveness of \OurMethod, we adopt two baselines: \textbf{1)} Standard prompt and \textbf{2)} Zero-shot CoT~\cite{kojima2022large}. 
The standard prompt only involves the spatial question. Zero-shot CoT is a task agnostic prompting technique and appends ``Let's think step by step'' afters the user query. The full prompt templates are shown in the Appendix~\ref{supp:full_text_prompt}.

\subsubsection{Main results}~\label{sec:main_results}
We evaluate the success rate $\delta_{\leq 2}$ in Table~\ref{tab:main_results}. For \OurMethod, we apply \OurMethodSimple on \gptv and \gpto and \OurMethodComplex on \gemini and \geminiflash. 
As shown before in Sec.~\ref{sec:analysis}, \gpto outperforms all the other VLMs out of the box. Zero-shot CoT significantly improves \gptv over standard prompt, but fail to boost the performances in \gemini and \geminiflash with only marginal or even negative improvements. 
On the contrary, \OurMethod improves the success rate across almost all VLMs, resulting in 47, 22, 30, and 2 points in \gemini, \geminiflash, \gptv, and \gpto, respectively. This drastically closes the gap between the \gpto and other VLMs. In Sec.~\ref{sec:analysis}, we observe a 41 point gap between \gpto and the second-best VLM in \benchmark when receive the standard prompt. With \OurMethod, the gap between \gpto and the second-best VLM is reduced to 13 points.
%Moreover, when \gemini is prompted with \OurMethod, it performs better than applying standard prompt on \gpto. \DA{this is only in one split? clarify?}

\subsubsection{Does \OurMethod encourage the use of references?}~\label{sec:frequency}
\begin{figure*}[t]
\includegraphics[width=\textwidth]{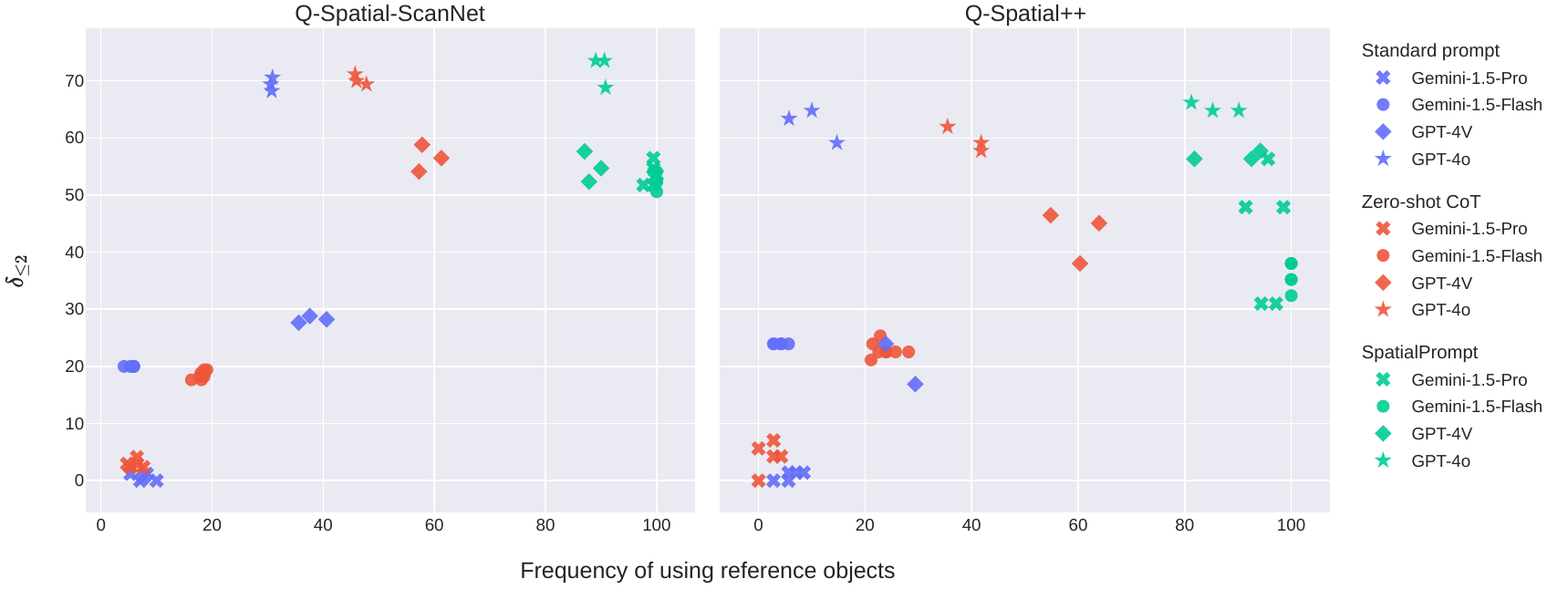}
\centering
\caption{\textbf{Success rates versus the frequencies of using reference objects.} 
\textcolor{plotly_green}{Green} corresponds to \texttt{SpatialPrompt}, \textcolor{plotly_red}{Red} corresponds to Zero-shot CoT, and \textcolor{plotly_blue}{Blue} corresponds to standard prompt.
Across all VLMs (represented by distinct symbols) and text prompts (indicated by distinct colors), the two variables show a strong correlation, with the values of 0.69 and 0.91 in \scannetqas and \inhouseqas, respectively.
This result is aligned with our hypothesis in Sec.~\ref{sec:why_gpt4o_is_good}:  \textit{Reasoning via a reference object directly improves overall accuracy on quantitative spatial questions}.}
\label{fig:reference_freq_scatter}
%\DA{Let's walk the reader on this figure? what the reader should look at? e.g. why the figure is important  why it is aligned with our hypoth}
\end{figure*}

The motivation behind \OurMethod is to encourage VLMs to use reference objects within the scene to answer the quantitative spatial questions. We, therefore, compare the frequency of whether a response involves reasoning via reference objects to answer the question. Same as the analysis in Sec.~\ref{sec:why_gpt4o_is_good}, we adopt a separate GPT-4o instance to determine if the model responses involves reference objects. 
Fig.~\ref{fig:reference_freq_scatter} visualizes the relations between the frequency of using any reference objects and the success rates $\delta_{\leq 2}$. The Spearman correlation coefficients of the two variables are 0.69 and 0.61 in \scannetqas and \inhouseqas, respectively. This empirically corroborates the hypothesis in Sec.~\ref{sec:why_gpt4o_is_good}.

% Figure.~\ref{fig:reference_freq_scatter} visualizes the relations between the the frequency of using references objects and $\delta_{\leq 2}$. The performance with identical prompts is shown in matching colors (blue, red, green), and results with the same VLMs are represented by consistent symbols. The figure is aligned with our hypothesis in Sec.~\ref{sec:analysis}: \textit{Reasoning via a reference object directly improves overall accuracy on quantitative spatial questions}. 

%In Table~\ref{tab:compare_reference_frequency}, we show the frequency of different VLMs and different text prompts. While zero-shot CoT increases the chances of involving reference objects in the responses, \OurMethod increases the most by explicitly asking the model to use the reference objects in the image. 
%\AL{explain}
%\DA{Maybe a correlation plot could make it nicer?}

\subsubsection{Open-sourced VLMs}\label{sec:open_source}

In this work we mainly focus on SoTA VLMs (\ie, commercial VLMs)  as larger models are known for their superior reasoning capabilities and their ability to follow instruction prompts more closely. 
Here, however, we also investigate the off-the-shelf performance and capabilities of open-source VLMs, \ie, LLaVA-v1.6-34b~\cite{liu2024llavanext}.

Table~\ref{tab:llava_results} shows the performances of LLaVA in \benchmark using standard prompt. Surprisingly, LLaVA achieves 60.59 in \scannetqas, outperforming most commercial VLMs including \gemini, \geminiflash, and \gptv. However, LLaVA falls more than 20 points when evaluated on \inhouseqas. Interestingly, when qualitatively analyzing the outputs of the models, we find that they essentially ``predict'' numbers rather than demonstrating any intuitive reasoning behavior, as seen in more powerful VLMs. We initially speculated that this might indicate the model was trained on some collection of datasets including ScanNet data. However, we found no reference suggesting that LLaVA directly uses ScanNet in its training dataset.

%Specifically, we evaluate the success rate in LLaVA-1.6-34b~\cite{liu2024llavanext} with different text prompts in Table~\ref{tab:lava_text_prompt}. We first observe that LLaVA achieves 60.59 in \scannetqas, nearly on par with GPT-4o (achieving 64.13) when receiving the standard prompt. However, LLaVA falls more than 20 points when evaluated on \inhouseqas. 
%Interestingly, when qualitatively analyzing the outputs of the models, we find that they essentially "predict" numbers rather than demonstrating any intuitive reasoning behavior, as seen in more powerful VLMs. We initially speculated that this might indicate the model was trained on some collection of datasets including ScanNet data. However, we found no reference suggesting that LLaVA directly uses ScanNet in its training dataset.

Additionally, we found LLaVA does not perform well when receiving zero-shot CoT prompts. When receiving zero-shot CoT prompts, the performances drop by over 20 and 4 points in \scannetqas and \inhouseqas. For \OurMethod, we adopt \OurMethodComplex, leading to a decrease around 9 points in \scannetqas. In sharp contrast, \OurMethodComplex improves 9 points in \inhouseqas.
We hypothesize several reasons for this: (i) LLaVA's capabilities are not as developed as those in very large commercial models, and inducing reasoning structures in-context via prompts  do not yet lead to the same level of improvements; (ii) LLaVA may fail to follow the instruction prompts effectively; (iii) LLaVA may not be proficient at visual comparison, so even if it increases the use of reference objects, it does not yield better performance.
This observation is aligned with LLMs in language tasks~\cite{Wei2022ChainOT,kojima2022large}, suggesting that the reasoning capabilities usually 
emerge when the model size exceeds 100B.

%Additionally, we found LLava does not performs well when receiving zero-shot CoT prompts. Specifically, the success rate drops by nearly 20 and 5 points in \scannetqas and \inhouseqas splits. \OurMethod leads to performance increases in \inhouseqas but fail to improve in \scannetqas. 
%We hypothesize several reasons for this: (i) LLaVA-1.6-34b's capabilities are not as developed as those in very large commercial models, and inducing reasoning structures in-context via prompts  do not yet lead to the same level of improvements; (ii) LLaVA may fail to follow the instruction prompts effectively; (iii) LLaVA may not be proficient at visual comparison, so even if it increases the use of reference objects, it does not yield better performance.
%\DA{Copy and paste the finding from CoT here about models requiring 100B or more to effectively use chain-of-thoughs.} \AL{Please review.}
%\DA{please check and revise as need it}

% \usepackage{booktabs}
% \usepackage{graphicx}
\begin{table}[t]
\centering
\resizebox{\columnwidth}{!}{%
\begin{tabular}{@{}lrr@{}}
\toprule
                & \scannetqas   & \inhouseqas   \\ \cmidrule(l){2-3} 
Standard prompt & \textbf{60.59} & 36.62            \\
Zero-shot CoT   &           40& 32.39          \\
\OurMethod      &           51.76& \textbf{45.07} \\ \bottomrule
\end{tabular}%
}
\caption{\textbf{Success rate $\delta_{\leq 2}$ of LLaVA in \scannetqas and \inhouseqas.} }%\AL{Not updated yet}}
%\vspace{-5mm}
\label{tab:llava_results}
\end{table}

\section{Limitations}\label{sec:limitations}

In this section, we discuss the limitations of our work from two aspects: the proposed benchmark and the proposed prompting technique.

\benchmark is limited by its size due to the time-consuming questions-answering annotation, especially the time spent to measure high-precision distances. In cases where the end points of the distance is ambiguous, the annotators are required to clicking around the nearby point cloud to ensure the measurement quality. For example, when annotating the shortest distance between two objects, to find the end points of the shortest distance can be time-consuming. As compared to prior works~\cite{spatial_vlm,spatial_rgpt}, our benchmark is around four times smaller. 

\OurMethod is essentially trading tokens to performances since the appending text prompt encourages VLMs to generate the reasoning paths in a detailed manner. This makes it less likely to be applied in the application requires real-time inference. Additionally, the motivation of \OurMethod is that we want to encourage VLMs to use the context information to improve quantitative spatial reasoning. However, when the image consists of a clean background, it is possible that there is little visual cues in the image; therefore, adding \OurMethod hurts or confuses VLMs. Finally, as shown in Sec.~\ref{sec:open_source}, \OurMethod results in inconsistent improvements for open-source VLMs. It is unclear that if the proposed \OurMethod only works when the model scale is large enough.

To summarize, \benchmark is mainly limited by its sizes due to the costly labeling process. On the other hand, \OurMethod can be limited by inference speed, image context and maybe model scales. 

\iffalse
\section{Broader impacts}
Quantitative spatial reasoning is itself challenging even for humans. The distance output from the VLMs can be very wrong and should not be used in any high-stake applications for now, such as autonomous driving.
\fi

\section{Conclusion}

In conclusion, our work introduces \benchmark, a novel benchmark tailored for evaluating VLMs on quantitative spatial reasoning and that incorporates 5 different  question categories. Through systematic analysis, we have identified the strengths and weaknesses of current SoTA VLMs in this task. Furthermore, we have developed a prompt-based technique that leverages naturally emerging reasoning paths to enhance VLM performance on quantitative spatial tasks. Our findings demonstrate that instructing VLMs to use reference objects in their reasoning pathways significantly boosts their performance without the need for additional data or fine-tuning.

\clearpage
\bibliography{main}

\clearpage
\appendix

\section{Table of Contents}
We organize the Appendix in the following structures:
\begin{enumerate}
    \item Sec.~\ref{supp:full_exp_results} provides the full experiment results of \gemini, \geminiflash, \gptv, and \gpto in Table~\ref{tab:delta_2_full_commercial}, Table~\ref{tab:delta_2_full_llava}, Table~\ref{tab:delta_1.25_full_commercial}.
    \item Sec.~\ref{supp:full_text_prompt} shows all the text prompts used in this work.
    \item Sec.~\ref{supp:qualitative_results} provide the qualitative results.
    \item Sec.~\ref{supp:implementation_details} shows the implementation details including generation parameters and computational resources used in this work.
    \item Sec.~\ref{supp:license} shows the license related to this work
\end{enumerate}

\section{Full experimental results}\label{supp:full_exp_results}

This section provides the additional experimental results:

%In the main paper, we demonstrate that we have two variants of \OurMethod: \OurMethodSimple and \OurMethodComplex. We use \OurMethodComplex for Gemini 1.5 Pro and Gemini 1.5 Flash and \OurMethodSimple for GPT-4V and GPT-4o, respectively. 

\textbf{Full version of Table~\ref{tab:main_results}.}
Table~\ref{tab:delta_2_full_commercial} shows the full results of Gemini 1.5 Pro, Gemini 1.5 Flash, GPT-4V, and GPT-4o along with their standard deviations. We find that Gemini 1.5 Pro only works when prompted with detail procedures and Gemini 1.5 Flash works pretty well when prompted with abstract and detailed procedures. On the other hand, prompting GPT-4V and GPT-4o with detail procedures hurts the performances in \benchmark. We hypothesize that since GPT-4V and GPT-4o are more competent models for quantitative spatial reasoning, constraining the reasoning path with detail procedures limits the performances.

\textbf{Full version of Table~\ref{tab:llava_results}.}
Table~\ref{tab:delta_2_full_llava} includes different variants of LLaVA~\cite{liu2024llavanext,liu2023improvedllava}. As show in the main paper, LLaVA-v1.6-34b demonstrate strong performances on \scannetqas, but falls more than 20 points in \inhouseqas. From Table~\ref{tab:delta_2_full_llava}, we find that different LLaVA variants has very different preferred prompts in \benchmark. 
These inconsistent preferences might result from the findings in the prior works~\cite{Wei2022ChainOT,kojima2022large} where models under 100B parameters exhibit unreliable reasoning paths.

\textbf{Evaluation in $\delta_{\leq 1.25}$.} 
Table~\ref{tab:delta_1.25_full_commercial} shows the performances in $\delta_{\leq 1.25}$. By comparing Table~\ref{tab:delta_2_full_commercial} that evaluates in $\delta_{\leq 2}$ and Table~\ref{tab:delta_1.25_full_commercial}, we find that the standard deviations generally increases when evaluated in $\delta_{\leq 1.25}$ due to a smaller threshold.

\textbf{Full table of the frequency of using reference objects for each VLM.}

In Table~\ref{tab:compare_reference_frequency}, we show the frequency of different VLMs and different text prompts. While zero-shot CoT increases the chances of involving reference objects in the responses, \OurMethod increases the most by explicitly asking the model to use the reference objects in the image.

%\textbf{2D scatter plot of the frequency of using references objects and $\delta_{\leq 2}$.} 
%Figure.~\ref{fig:reference_freq_scatter} visualizes the relations between the the frequency of using references objects and $\delta_{\leq 2}$. The performance with identical prompts is shown in matching colors (blue, red, green), and results with the same VLMs are represented by consistent symbols. The figure is aligned with our hypothesis in Sec.~\ref{sec:analysis}: \textit{Reasoning via a reference object directly improves overall accuracy on quantitative spatial questions}. 

\textbf{Human evaluation.} To gauge the difficulty of the introduced benchmark, \benchmark, we ask three humans to perform quantitative spatial reasoning. Each human answer a sampled set of 40 quantitative spatial reasoning questions randomly draw from both splits. Table~\ref{tab:human_baselines} show the success rate $\delta_{\leq 2}$ of 3 humans. Though GPT-4o demonstrates strong performances as compared to other VLMs, it still talls behind humans by more than 30 points. This suggests that the curated benchmark is easy to humans but difficult to VLMs.

% \usepackage{booktabs}
% \usepackage{graphicx}
\begin{table*}[t]
\centering
\resizebox{\textwidth}{!}{%
\begin{tabular}{@{}lrrrrrrrr@{}}
\toprule
 &
  \multicolumn{4}{c}{\scannetqas} &
  \multicolumn{4}{c}{\inhouseqas} \\
 &
  \multicolumn{1}{c}{Gemini 1.5 Pro} &
  \multicolumn{1}{c}{Gemini 1.5 Flash} &
  \multicolumn{1}{c}{GPT-4V} &
  \multicolumn{1}{c}{GPT-4o} &
  \multicolumn{1}{c}{Gemini 1.5 Pro} &
  \multicolumn{1}{c}{Gemini 1.5 Flash} &
  \multicolumn{1}{c}{GPT-4V} &
  \multicolumn{1}{c}{GPT-4o} \\ \cmidrule(l){2-9} 
Standard prompt &
  0.59 {\small $\pm$ 0.52} &
  20 {\small $\pm$ 0.} &
  28.24 {\small $\pm$ 0.48} &
  69.41 {\small $\pm$ 0.96} &
  0.99 {\small $\pm$ 0.62}&
  23.73 {\small $\pm$ 0.}&
  18.81 {\small $\pm$ 1.98}&
  61.06 {\small $\pm$ 1.86}\\
Zero-shot CoT &
  2.94 {\small $\pm$ 0.64} &
  18.53 {\small $\pm$ 0.65} &
  \textbf{56.47 {\small $\pm$ 1.92}} &
  70.2 {\small $\pm$ 0.7} &
  4.16 {\small $\pm$ 1.91}&
  26.11 {\small $\pm$ 0.68}&
  46.53 {\small $\pm$ 2.8}&
  59.74 {\small $\pm$ 1.23}\\
\OurMethodSimple &
  6.47 {\small $\pm$ 1.17} &
  49.76 {\small $\pm$ 1.8} &
  54.9 {\small $\pm 2.16$} &
  \textbf{71.96 {\small $\pm$ 2.21}} &
  5.35 {\small $\pm$ 1.05} &
  \textbf{55.77 {\small $\pm$ 2.45}} &
  \textbf{53.47 {\small $\pm$ 3.23}}&
  \textbf{62.71 {\small $\pm$ 1.68}}\\
\OurMethodComplex &
  \textbf{53.65 {\small $\pm$ 1.79}} &
  \textbf{52.71 {\small $\pm$ 1.32}} &
  53.14 {\small $\pm$ 0.55} &
  63.33 {\small $\pm$ 1.54} &
  \textbf{43.17 {\small $\pm$ 7.21}}&
  38.42 {\small $\pm$ 2.37}&
  35.68 {\small $\pm$ 1.75} &
  58.69 {\small $\pm$ 1.75} \\ \bottomrule
\end{tabular}%
}
\caption{\textbf{Full table of the success rate $\delta_{\leq 2}$ of Gemini 1.5 Pro, Gemini 1.5 Flash, GPT-4V, and GPT-4o.} All numbers are averaged over 5 different runs,  except for GPT-4V and GPT-4o,
which are run on three seeds. Each number is followed by their standard deviations.}
\label{tab:delta_2_full_commercial}
\end{table*}
% \usepackage{booktabs}
% \usepackage{graphicx}
\begin{table*}[t]
\centering
\resizebox{\textwidth}{!}{%
\begin{tabular}{@{}lrrrrrrrr@{}}
\toprule
 &
  \multicolumn{4}{c}{\scannetqas} &
  \multicolumn{4}{c}{\inhouseqas} \\
 &
  \multicolumn{1}{c}{LLaVA-v1.6-34b} &
  \multicolumn{1}{c}{LLaVA-v1.6-13b} &
  \multicolumn{1}{c}{LLaVA-v1.5-13b} &
  \multicolumn{1}{c}{LLaVA-v1.5-7b} &
  \multicolumn{1}{c}{LLaVA-v1.6-34b} &
  \multicolumn{1}{c}{LLaVA-v1.6-13b} &
  \multicolumn{1}{c}{LLaVA-v1.5-13b} &
  \multicolumn{1}{c}{LLaVA-v1.5-7b} \\ \cmidrule(l){2-9} 
Standard prompt &
  \textbf{60.59}&
  55.88&
  42.24&
  40&
  36.62&
  38.03&
  35.21&
  29.58\\
Zero-shot CoT &
   40&
  \textbf{58.24}&
  41.76&
  44.12&
  32.39&
  \textbf{47.89}&
  23.94&
  \textbf{35.21}\\
\OurMethodSimple &
  47.65&
  44.12&
  35.29&
  47.06&
  39.44&
  45.07&
  35.21&
  \textbf{35.21}\\
\OurMethodComplex &
  51.76&
  35.29&
   \textbf{44.71}&
  \textbf{49.41}&
  \textbf{45.07} &
  39.15&
  \textbf{46.68} &
  33.8\\ \bottomrule
\end{tabular}%
}
\caption{\textbf{Full table of the success rate $\delta_{\leq 2}$ of LLaVA at different versions and model sizes.} All numbers are averaged over 5 different runs and followed by their standard deviations.}
\label{tab:delta_2_full_llava}
\end{table*}
% \usepackage{booktabs}
% \usepackage{graphicx}
\begin{table*}[t]
\centering
\resizebox{\textwidth}{!}{%
\begin{tabular}{@{}lrrrrrrrr@{}}
\toprule
 &
  \multicolumn{4}{c}{\scannetqas} &
  \multicolumn{4}{c}{\inhouseqas} \\
 &
  \multicolumn{1}{c}{Gemini 1.5 Pro} &
  \multicolumn{1}{c}{Gemini 1.5 Flash} &
  \multicolumn{1}{c}{GPT-4V} &
  \multicolumn{1}{c}{GPT-4o} &
  \multicolumn{1}{c}{Gemini 1.5 Pro} &
  \multicolumn{1}{c}{Gemini 1.5 Flash} &
  \multicolumn{1}{c}{GPT-4V} &
  \multicolumn{1}{c}{GPT-4o} \\ \cmidrule(l){2-9} 
Standard prompt &
  0.35 {\small $\pm$ 0.28}&
  10.59 {\small $\pm$ 0.}&
  13.92 {\small $\pm$ 0.99}&
  32.55 {\small $\pm$ 1.2}&
  0. {\small $\pm$ 0.}&
  16.83 {\small $\pm$ 0.}&
  8.91 {\small $\pm$ 0.99}&
  22.77 {\small $\pm$ 3.7}\\
Zero-shot CoT &
  1.41 {\small $\pm$ 0.47}&
  7.06 {\small $\pm$ 0.58}&
  \textbf{23.53 {\small $\pm$ 1.73}}&
  33.53 {\small $\pm$ 0.83}&
  0.4 {\small $\pm$ 0.48}&
  11.51 {\small $\pm$ 0.84}&
  20.79 {\small $\pm$ 2.13}&
  20.46 {\small $\pm$ 3.26}\\
\OurMethodSimple &
  3.41 {\small $\pm$ 0.86}&
  18.47 {\small $\pm$ 0.95}&
  22.94 {\small $\pm$ 0.48}&
  \textbf{35.69 {\small $\pm$ 1.99}}&
  1.41 {\small $\pm$ 1.25}&
  \textbf{22.25 {\small $\pm$ 2.25}}&
  \textbf{22.44 {\small $\pm$ 5.13}}&
  21.78 {\small $\pm$ 4.2}\\
\OurMethodComplex &
  \textbf{38 {\small $\pm$ 0.59}}&
  \textbf{21.29 {\small $\pm$ 1.76}}&
  19.8 {\small $\pm$ 2.46}&
  26.67 {\small $\pm$ 2.26}&
  \textbf{12.08 {\small $\pm$ 3.33}}&
  12.67 {\small $\pm$ 1.58}&
  13.15 {\small $\pm$ 3.31}&
  \textbf{27.7 {\small $\pm$ 2.65}} \\ \bottomrule
\end{tabular}%
}
\caption{\textbf{Full table of the success rate $\delta_{\leq 1.25}$ of Gemini 1.5 Pro, Gemini 1.5 Flash, GPT-4V, and GPT-4o.} All numbers are averaged over 5 different runs,  except for GPT-4V and GPT-4o,
which are run on three seeds. Each number is followed by their standard deviations.}
\label{tab:delta_1.25_full_commercial}
\end{table*}

\begin{table*}[t]
\centering
\resizebox{\textwidth}{!}{%
\begin{tabular}{@{}lrrrrrrrr@{}}
\toprule
 &
  \multicolumn{4}{c}{\scannetqas} &
  \multicolumn{4}{c}{\inhouseqas} \\
 &
  \multicolumn{1}{c}{Gemini 1.5 Pro} &
  \multicolumn{1}{c}{Gemini 1.5 Flash} &
  \multicolumn{1}{c}{GPT-4V} &
  \multicolumn{1}{c}{GPT-4o} &
  \multicolumn{1}{c}{Gemini 1.5 Pro} &
  \multicolumn{1}{c}{Gemini 1.5 Flash} &
  \multicolumn{1}{c}{GPT-4V} &
  \multicolumn{1}{c}{GPT-4o} \\ \cmidrule(l){2-9} 
Standard prompt &
  7.64&
  5.42&
  37.9&
  30.7&
  5.91&
  3.94&
  26.64&
  10.14\\
Zero-shot CoT &
  6.11&
  18.12&
  58.75&
  46.48&
  1.97&
  27.71&
  59.7&
  39.68\\
\OurMethod &
  \textbf{99.17}&
  \textbf{99.8}&
  \textbf{88.28}&
  \textbf{90.14}&
  \textbf{95.42}&
  \textbf{100}&
  \textbf{89.51}&
  \textbf{85.55}\\ \bottomrule
\end{tabular}%
}
\caption{\textbf{Frequency of whether the responses involve using reference objects of different VLMs and prompting techniques.} The proposed prompt \OurMethod consistently lead to higher chances to have reference objects involved in the responses. }
\label{tab:compare_reference_frequency}
\end{table*}

\begin{table}[t]
\centering
\resizebox{\columnwidth}{!}{%
\begin{tabular}{@{}lr@{}}
\toprule
           & Sampled Q-Spatial-Bench \\ \cmidrule(l){2-2} 
Human A    & 90                      \\
Human B    & 82.5                    \\
Human C    & 97.5                    \\
Human Avg. & 90                      \\ \bottomrule
\end{tabular}%
}
\caption{\textbf{Human performances in \benchmark.} Though GPT-4o demonstrates strong performances as compared to other VLMs, it still falls behind humans by more than 30 points. This suggests that the curated benchmark is easy to humans but difficulty to VLMs.}
\label{tab:human_baselines}
\end{table}

\section{Full text prompt}\label{supp:full_text_prompt}
This section provides the full text prompts used in our work, 

\begin{enumerate}
    \item Fig.~\ref{fig:reference_extraction_prompt}: The text prompt used to determine whether the VLM responses involves using any reference objects. It is used in Table~\ref{tab:gpt4o_analysis_contingency_table}, Table~\ref{tab:gpt4o_analysis}, and Table~\ref{tab:compare_reference_frequency}.
    \item Fig.~\ref{fig:standard_prompt}: The text prompt is used in Table~\ref{tab:preliminary_analysis} and Table~\ref{tab:vlm_strength_weakness}. It is referred to as ``Standard prompt'' in Sec.~\ref{sec:spatial_prompt}.
    \item Fig.~\ref{fig:llava_standard_prompt}: The text prompt is slightly modified from the prompt in Fig.~\ref{fig:standard_prompt} since LLaVA~\cite{liu2024llavanext} constantly fails to follow the right output format. Following prior work~\cite{kojima2022large}, we adopt a two-stage generation strategy to encourage the correct formats in the responses.
    \item Fig.~\ref{fig:spatial_prompt_steps}: The text prompt is referred to as \OurMethodComplex in Sec.~\ref{sec:spatial_prompt}. 
    \item Fig.~\ref{fig:llama3_qa_gen}: The text prompt is used to convert a triplet of an image, a distance, and a description of the distance to a question-answering pair. 
\end{enumerate}

\iffalse
Fig.~\ref{fig:_reference_extraction_prompt} shows the text prompt used to determine whether the VLM responses involves using any reference objects which is used in Table~\ref{tab:gpt4o_analysis_contingency_table}, Table~\ref{tab:gpt4o_analysis}, and Table~\ref{tab:compare_reference_frequency}. Fig.~\ref{fig:_stanard_prompt} shows the standard prompt used in Sec.~\ref{sec:analysis} and Sec.~\ref{sec:main_results}. Fig.~\ref{fig:_spatial_prompt_steps} shows full details in \OurMethodComplex. Fig.~\ref{fig:_system_prompt} shows the system prompt used in Table~\ref{tab:vlm_strength_weakness}, Table~\ref{tab:preliminary_analysis}, Table~\ref{tab:main_results}, and Table~\ref{tab:lava_text_prompt}.

For Sec.~\ref{sec:benchmark}, we use Llama3-70b to convert the triplet of an image, a distance, and a description to the distance to a question-answering pair. Fig.~\ref{fig:_llaba3_qa_gen} show the full prompt.
\fi

\section{Qualitative results}\label{supp:qualitative_results}

\textbf{GPT-4V responses with different text prompts.} In Fig.~\ref{supp:qualitative_res_gpt4v}, we compare GPT-4V responses with different text prompts. When using standard prompt, GPT-4V tends to perform a very rough estimate directly. On the other hand, when prompted with zero-shot CoT and \texttt{SpatialPrompt}, GPT-4V leverages surrounding objects for quantitative spatial reasoning. 

\textbf{Gemini 1.5 Pro responses with different text prompts.} In Fig.~\ref{supp:qualitative_res_geminipro}, we compare Gemini 1.5 Pro responses with different text prompts. Among all the evaluated prompts, SpatialPropmt is the only one that successfully elicit quantitative spatial reasoning in Gemini-1.5-Pro.

%\textbf{Qualitative results across different VLMs.}
%In Fig.~\ref{supp:all_vlms_qualitative_res}, we compare the resposnes from different VLMs.

%\textbf{Qualitative results of LLaVA-1.6-34b.} Fig.~\ref{supp:llava_qualitative_res} shows the sampled responses from LLaVA-1.6-34b.

\textbf{Failure cases of GPT-4o.} Fig.~\ref{supp:qualitative_res_gpt4o_fail} shows the common failure case of GPT-4o. We find that GPT-4o have a tendency to use floor tiles as references, leading to inaccurate estimation at times.

\section{Implementation details}\label{supp:implementation_details}
In this work, we employ deterministic sampling and conduct each experiment using five different random seeds, except for \gptv and \gpto, which are run on three seeds.
The specific endpoint for \gpto is \texttt{gpt-4o-2024-05-13}, for \gptv is \texttt{gpt-4-turbo-2024-04-09}, for \gemini is \texttt{gemini-1.5-pro-001}, and for \geminiflash is \texttt{gemini-1.5-flash-001}.
All GPT-4o, GPT-4V, and Gemini experiments are run with the provided API.

It is worth noted that most VLM APIs do not support fully reproducible results. That is, even when the temperature is set of $0$, there is still randomness in the results. We empirically find that setting top\_p to a very small value $1^{-14}$ and temperature to $1$ produces better than setting top\_p to $1$ and temperature to $0$. For all experiment, we set  top\_p to $1^{-14}$ and temperature to $1$.

For LLaVA experiments, we set temperature to $0$ and top\_p to $1.$.  For LLaVA, we run on a machine with 2 A100 GPUs.

For the logistic regression analysis, we use the default parameters in statsmodels~\cite{seabold2010statsmodels}.

\begin{figure*}[t]
\centering  
\begin{tcolorbox}
\begin{Verbatim}[breaklines=true, commandchars=\\\{\}]
System: You are an AI assistant evaluating responses to questions about measuring distances in 3D space using 2D images. Check if the response uses reference objects or scales to answer the question. Answer "YES" if it does, "NO" if it does not, and "UNSURE" if uncertain. Begin your answer with "YES", "NO", or "UNSURE".
User: Question: \textcolor{red}{[SPATIAL QUESTION]}
Response: \textcolor{red}{[VLM RESPONSE]}
Assistant: \textcolor{red}{[OUTPUT]}
\end{Verbatim}
\end{tcolorbox}
\caption{\textbf{Text prompt for determining whether the VLM responses involves using any reference objects.} The text prompt is used in Table~\ref{tab:gpt4o_analysis_contingency_table}, Table~\ref{tab:gpt4o_analysis}, and Table~\ref{tab:compare_reference_frequency}.}\label{fig:reference_extraction_prompt}
\end{figure*}

\begin{figure*}[t]
\centering   
\begin{tcolorbox}
\begin{Verbatim}[breaklines=true, commandchars=\\\{\}]
System: You will be provided with a question and a 2D image. The question involves measuring the precise distance in 3D space through a 2D image. You will answer the question by providing a numeric answer consisting of a scalar and a distance unit in the format of """\textbackslash{}scalar\{scalar\} \textbackslash{}distance_unit\{distance unit\}""" at the end of your response.
User: Question: \textcolor{red}{[SPATIAL QUESTION]}
Assistant: \textcolor{red}{[OUTPUT]}
\end{Verbatim}
\end{tcolorbox}
\caption{\textbf{Standard prompt for quantitative spatial reasoning.} The text prompt is used in Table~\ref{tab:preliminary_analysis} and Table~\ref{tab:vlm_strength_weakness}. The prompt is referred to as ``Standard prompt'' in Section~\ref{sec:spatial_prompt}.}\label{fig:standard_prompt}
\end{figure*}

\begin{figure*}[t]
\centering   
\begin{tcolorbox}
\begin{Verbatim}[breaklines=true, commandchars=\\\{\}]
System: You will be provided with a question and a 2D image. The question involves measuring the precise distance in 3D space through a 2D image. You will answer the question by providing a numeric answer consisting of a scalar and a distance unit in the format of """\textbackslash{}scalar\{scalar\} \textbackslash{}distance_unit\{distance unit\}""" at the end of your response.
User: Question: \textcolor{red}{[SPATIAL QUESTION]}
Answer by providing a numeric answer consisting of a scalar and a distance unit in the format of """\textbackslash{}scalar\{scalar\} \textbackslash{}distance_unit\{distance unit\}""" at the end of your response.
Assistant: \textcolor{red}{[FIRST STAGE OUTPUT]} In conclusion, the final answer in the specified format is: """\textbackslash{}scalar\{\textcolor{red}{[SECOND STAGE OUTPUT]}
\end{Verbatim}
\end{tcolorbox}
\caption{\textbf{Standard prompt for LLaVA.} The text prompt is slightly modified from the prompt in Fig.~\ref{fig:standard_prompt} since LLaVA~\cite{liu2024llavanext} constantly fails to follow the right output format. Following prior work~\cite{kojima2022large}, we adopt a two-stage generation strategy to encourage the correct formats in the responses.}\label{fig:llava_standard_prompt}
\end{figure*}
\begin{figure*}[t]
\centering   
\begin{tcolorbox}
\begin{Verbatim}[breaklines=true, commandchars=\\\{\}]
System: You will be provided with a question and a 2D image. The question involves measuring the precise distance in 3D space through a 2D image. You will answer the question by providing a numeric answer consisting of a scalar and a distance unit in the format of """\textbackslash{}scalar\{scalar\} \textbackslash{}distance_unit\{distance unit\}""" at the end of your response.
User: Question: \textcolor{red}{[SPATIAL QUESTION]}

---

Use the following 4 steps sequentially to answer the question:

Step 1 **Analyze the question**

Step 2 **Identify up to 10 reference scales in the image, ranging from large to small sizes, and list them in the specified format**
- A reference scale must be typical in size.
- A reference scale can be the dimensions of an object or an object part.
- A reference scale must NOT be floor tiles or floor planks.
- Formulate the reference scales using the format: """The [choose from front-to-back, side-to-side, left-to-right, diameter, height (top to bottom edge), or mounting height (bottom edge to floor)] of [object or object part] is approximately [dimension estimate]."""

Step 3 **Propose a robust step-by-step plan to answer the question by using the reference scales in Step 2**
- A robust step-by-step plan performs the estimation in a coarse-to-fine manner.
  - First, use a reliable and large-sized reference scale as the primary reference for estimation.
  - Then, gradually use a reliable and smaller-sized reference scale for adjustment.
  - Repeat until the estimation is precise enough.
- When performing visual comparison, be aware of perspective distortion.
- Do NOT rely on pixel measurements from the images.

Step 4 **Focus on the image and follow the plan in Step 3 to answer the question**
Assistant: \textcolor{red}{[OUTPUT]}
\end{Verbatim}
\end{tcolorbox}
\caption{\textbf{Text prompt in \OurMethodComplex}. The text prompt is used in Sec.~\ref{sec:spatial_prompt}}\label{fig:spatial_prompt_steps}
\end{figure*}

\begin{figure*}[t]
\centering   
\begin{tcolorbox}
\begin{Verbatim}[breaklines=true, commandchars=\\\{\}]
User: You are an intelligent agent that can help me. You will be given a distance or a surface area spcified in an image and based on this information, you need to generate a visual question answer pair.

For example, 
User input: ("the minimum distance between the red mug and the blue fork", "30 cm")
Output: ("What is the minimum distance between the red mug and the blue fork in the image? Answer in centimeters.", "30")

User input: ("the available surface area on the floor", "400 sqft")
Output: ("What is the available surface area on the floor? Answer in sqft", "400")

User input: ("the available surface area on the kitchen counter", "2 meters in length and 40 cm in depth")
Output: ("What is the available surface area on the kitchen counter? Answer by square meters", "0.8 square meter")

Now it's your turn,
User input: ("\textcolor{red}{[DISTANCE DESCRIPTION]}", "\textcolor{red}{[DISTANCE VALUE]}")
Output: 
Assistant: \textcolor{red}{[OUTPUT]}
\end{Verbatim}
\end{tcolorbox}
\caption{\textbf{Text prompt used to convert an annotated triplet (image, distance, description) to a question-answering pair.} The text prompt is used to convert a triplet of an image, a distance, and a description of the distance to a question-answering pair. It is used in Sec.~\ref{sec:benchmark}.}\label{fig:llama3_qa_gen}
\end{figure*}

\begin{figure*}[t]
    \centering
    % First subfigure
    \begin{minipage}[t]{0.32\textwidth}
        \centering
        \includegraphics[width=\textwidth]{qualitative_results/example_images/qspatial_plus-057.jpeg}
        \caption*{(A)} % Annotating as A
    \end{minipage}
    % Second subfigure
    \begin{minipage}[t]{0.32\textwidth}
        \centering
        \includegraphics[width=\textwidth]{qualitative_results/example_images/qspatial_plus-018.jpeg}
        \caption*{(B)} % Annotating as B
    \end{minipage}
    % Third subfigure
    \begin{minipage}[t]{0.32\textwidth}
        \centering
        \includegraphics[width=\textwidth]{qualitative_results/example_images/qspatial_plus-030.jpeg}
        \caption*{(C)} % Annotating as C
    \end{minipage}
    \caption{\textbf{Sampled images from Q-Spatial++}}
    \label{fig:sample_images}
\end{figure*}

\begin{figure*}[t]
\centering
\begin{tcolorbox}%[enhanced, breakable]
% Add the image inside the tcolorbox
%\centering
%\includegraphics[width=0.2\textwidth]{latex/qualitative_results/example_images/qspatial_plus-057.jpeg} 
% Adjust the width as needed
%\vspace{5mm} % Optional: add some space between the image and the text
% Add the text below the image
\begin{Verbatim}[breaklines=true, commandchars=\\\{\}]
\textcolor{blue}{[Image A in Figure 10]}
\textcolor{blue}{Question: What is the minimum distance between the two orange chairs next to the window?}
\textcolor{blue}{Ground truth: 47 centimeters}

# Response with Standard prompt
Assistant: The minimum distance between the two orange chairs next to the window appears to be very small, possibly around a few centimeters, as they are placed quite close to each other. However, without a reference for scale or the ability to measure directly, it's difficult to provide an exact measurement. Based on visual estimation, the distance might be around (\textbackslash{}scalar\{5\} \textbackslash{}distance\_unit\{cm\}).

# Response with Zero-shot CoT
Assistant: To estimate the minimum distance between the two orange chairs next to the window, we can use visual cues from the image. The chairs are positioned close to each other, with a small table between them. The table appears to be narrow, suggesting that the chairs are not far apart.

Given the typical width of such small tables, which is usually around 50 to 60 centimeters, and considering the slight gap between the edge of the table and the chairs, we can estimate the minimum distance between the chairs. Assuming each gap is about 5 to 10 centimeters, the total distance between the two chairs, including the table and gaps, would likely be around 60 to 80 centimeters.

Therefore, the minimum distance between the two orange chairs is approximately (\textbackslash{}scalar\{70\} \textbackslash{}distance\_unit\{cm\}).

# Response with SpatialPrompt
Assistant: To estimate the minimum distance between the two orange chairs next to the window, we can use the black table between them as a reference object. The table appears to be a standard small side table, which typically measures around 50 cm in diameter.

Observing the image, the table fits snugly between the two chairs, suggesting that the distance between the chairs is slightly more than the diameter of the table, to accommodate the table without pressing against the chairs. Therefore, we can estimate the minimum distance between the chairs to be slightly more than 50 cm.

Considering the visual information and typical furniture dimensions, the minimum distance between the two orange chairs is approximately (\textbackslash{}scalar\{55\} \textbackslash{}distance\_unit\{cm\}).
\end{Verbatim}
\end{tcolorbox}
\caption{\textbf{GPT-4V responses with different text prompts.}}
\label{supp:qualitative_res_gpt4v}
\end{figure*}

\begin{figure*}[t]
\centering
\begin{tcolorbox}%[enhanced, breakable]
% Add the image inside the tcolorbox
%\centering
%\includegraphics[width=0.2\textwidth]{latex/qualitative_results/example_images/qspatial_plus-018.jpeg} 
% Adjust the width as needed
%\vspace{5mm} % Optional: add some space between the image and the text
% Add the text below the image
\begin{Verbatim}[breaklines=true, commandchars=\\\{\}]
\textcolor{blue}{[Image B in Figure 10]}
\textcolor{blue}{Question: What is the minimum distance between the coffee grinder and the blue cloth in the image?}
\textcolor{blue}{Ground truth: 15 centimeters}

# Reponse with Standard prompt
It is impossible to determine the distance between the coffee grinder and the blue cloth from a 2D image.  A 2D image does not provide depth information.  \textbackslash{}scalar\{\} \textbackslash{}distance_unit\{\}

# Reponse with Zero-shot CoT
It is impossible to measure the distance in 3D space from a 2D image.  We do not know how far away the camera is from the coffee grinder and the blue cloth.  We also do not know the angle at which the picture was taken.  All of these factors would affect the perceived distance between the two objects.  Without more information, we cannot answer this question. 

# Response with SpatialPrompt
Step 1 **Analyze the question**

We need to find the shortest distance between the coffee grinder and the blue cloth. This is the perpendicular distance from the front edge of the grinder's base to the edge of the cloth.

Step 2 **Identify up to 10 reference scales in the image, ranging from large to small sizes, and list them in the specified format**

1. """The height (top to bottom edge) of the coffee grinder is approximately 10 inches."""
2. """The diameter of the kettle's base is approximately 8 inches."""
\textcolor{gray}{[text ignored]}

Step 3 **Propose a robust step-by-step plan to answer the question by using the reference scales in Step 2**
\textcolor{gray}{[text ignored]}

Step 4 **Focus on the image and follow the plan in Step 3 to answer the question**

1. The distance between the coffee grinder and the cloth appears to be slightly less than half the height of the coffee grinder. This suggests an initial estimate of around 4-5 inches.
2. Comparing the distance to the kettle's base diameter, it seems closer to 4 inches than 5 inches.

Therefore, the minimum distance between the coffee grinder and the blue cloth is approximately  \textbackslash{}scalar\{4\} \textbackslash{}distance_unit\{inches\}. 
\end{Verbatim}
\end{tcolorbox}
\caption{\textbf{Gemini 1.5 Pro responses with different text prompts.}}
\label{supp:qualitative_res_geminipro}
\end{figure*}

\begin{figure*}[t]
\centering
\begin{tcolorbox}%[enhanced, breakable]
% Add the image inside the tcolorbox
%\centering
%\includegraphics[width=0.2\textwidth]{latex/qualitative_results/example_images/qspatial_plus-030.jpeg} 
% Adjust the width as needed
%\vspace{5mm} % Optional: add some space between the image and the text
% Add the text below the image
\begin{Verbatim}[breaklines=true, commandchars=\\\{\}]
\textcolor{blue}{[Image C in Figure 10]}
\textcolor{blue}{Question:What is the minimum distance between the garbage bin and the stool in the image?}
\textcolor{blue}{Ground truth: 7 centimeters}

# Reponse with Standard prompt
To determine the minimum distance between the garbage bin and the stool in the image, I will estimate based on the visual perspective and relative sizes of the objects.

The garbage bin and the stool appear to be relatively close to each other. Given the perspective and the floor tiles as a reference, the minimum distance between the garbage bin and the stool is approximately:

\textbackslash{}scalar\{30\} \textbackslash{}distance_unit\{cm\}
\end{Verbatim}
\end{tcolorbox}
\caption{\textbf{Common failure cases of GPT-4o}}
\label{supp:qualitative_res_gpt4o_fail}
\end{figure*}

\section{License}\label{supp:license}
In \scannetqas, we use ScanNet images under the license at 
\url{https://kaldir.vc.in.tum.de/scannet/ScanNet_TOS.pdf}

\iffalse
\subsection{Prompt Templates}\label{supp:prompt_template}
Fig.~\ref{fig:gemini_prompts} shows the prompt for Gemini-1.5-pro

\input{prompts/gemini_prompts}
\input{prompts/gpt4o_prompts}

\AL{
To include
\begin{enumerate}
    \item Prompt to determine if the responses contain reference objects
    \item Llama3 prompt to generate QAs
    \item Histogram of answer in different splits
    \item Sampling parameters
\end{enumerate}
}
\fi

\end{document}